\documentclass[conference]{IEEEtran}
\IEEEoverridecommandlockouts
\usepackage{cite}
\usepackage{amsmath,amssymb,amsfonts}
\usepackage{algorithmic}
\usepackage{graphicx}
\usepackage{textcomp}
\usepackage{xcolor}
\usepackage{times}
\usepackage{epsfig}
\usepackage{booktabs}
\usepackage{multirow}
\usepackage{float}

\def\BibTeX{{\rm B\kern-.05em{\sc i\kern-.025em b}\kern-.08em
    T\kern-.1667em\lower.7ex\hbox{E}\kern-.125emX}}
    
\begin{document}

\title{
MicroExpNet: An Extremely Small and Fast Model For Expression Recognition From Face Images
}

\author{\IEEEauthorblockN{
Ilke Cugu, 
Eren Sener,
Emre Akbas}
\IEEEauthorblockA{\textit{Department of Computer Engineering}, 
\textit{Middle East Technical University}, 
Ankara, Turkey \\
\texttt{\{cugu.ilke,sener.eren,eakbas\}@metu.edu.tr}}
}

\IEEEoverridecommandlockouts
\IEEEpubid{\makebox[\columnwidth]{978-1-7281-3975-3/19/\$31.00~\copyright2019 IEEE
\hfill} \hspace{\columnsep}\makebox[\columnwidth]{ }}
\maketitle
\IEEEpubidadjcol 

\begin{abstract}
This paper is aimed at creating extremely small and fast convolutional neural networks (CNN) for the problem of facial expression recognition (FER) from frontal face images. To this end, we employed the popular knowledge distillation (KD) method and identified two major shortcomings with its use: 1) a fine-grained grid search is needed for tuning the temperature hyperparameter and 2) to find the optimal size-accuracy balance, one needs to search for the final network size (or the compression rate). On the other hand, KD is proved to be useful for model compression for the FER problem, and we discovered that its effects get more and more significant with decreasing model size. In addition, 
we hypothesized that translation invariance achieved using max-pooling layers would not be useful for the FER problem as the expressions are sensitive to small, pixel-wise changes around the eye and the mouth. However, we have found an intriguing improvement in generalization when max-pooling is used. We conducted experiments on two widely-used FER datasets, CK+ and Oulu-CASIA. Our smallest model (MicroExpNet), obtained using knowledge distillation, is less than $1$MB in size and works at $1851$ frames per second on an Intel i7 CPU. Despite being less accurate than the state-of-the-art, MicroExpNet still provides significant insights for designing a microarchitecture for the FER problem.
\end{abstract}


\section{Introduction}
\label{intro}

Expression recognition from frontal face images is an important aspect of human-computer interaction and has many potential applications, especially in mobile devices. Face detection models have long been deployed in mobile devices, and relatively recently, face recognition models are also being used, e.g. for face-based authentication. Arguably, one of the next steps is the mobile deployment of facial expression recognition models. Therefore, creating small and fast models is an important goal. In order to asses the current situation, we looked at the size and runtime speeds of two representatives, current state-of-the-art models, namely PPDN \cite{zhao2016peak} and FN2EN \cite{ding2017facenet2expnet}. In terms of the number of total parameters in the network, both models are in the order of millions (PPDN has 6M and FN2EN has 11M). In terms of speed, both models run at $9-11$ms per image on a GTX $1050$ GPU, however, on an Intel i7 CPU, while PPDN takes $57.18$ ms, FN2EN takes $96.08$ ms (further details in Table \ref{tab:modelSizesAndSpeeds}).

The central question that motivated the present work was how much we could push the size and speed limits so that we end up with a compact expression recognition model that still works reasonably well. To this end, we focused only on frontal face images and first explored training a large model on two widely used benchmarks FER datasets, CK+~\cite{lucey2010extended} and Oulu-Casia~\cite{zhao2011facial}, by using the Inception\_v3~\cite{szegedy2016rethinking} model. Then, using the ``knowledge distillation'' (KD) method~\cite{hinton2015distilling}, we created a family of small and fast models. In the KD method, there is a large, cumbersome model called the \emph{teacher} (Inception\_v3 in our case) and a relatively much smaller model called the \emph{student}. The student is trained to ``mimic'' the softmax values of the teacher via a \emph{temperature} hyperparameter (see Eq. \ref{lossFunc}). We have experimented on four student networks with different sizes. The smallest one, called \textbf{MicroExpNet}, is $100$x smaller in size and has $335$x fewer parameters compared to the teacher.

We found two major shortcomings of the KD method. First, the temperature hyperparameter does not seem to have any meaningful relationship with the accuracy of the student model. We found that the accuracy fluctuates between low and high values as the temperature is swept across a wide range. In order to find a high-accuracy temperature, one needs to do a fine-grained grid search. Second, in the KD method, the final student model size (i.e. the compression rate) is given as input. Therefore, to find the optimal size-accuracy balance, one needs to search for the size, too. 

We also hypothesized that invariance to translation achieved using max-pooling layers would not be useful for the FER problem as the expressions are sensitive to small, pixel-wise changes around the eye and the mouth. However, we empirically found that this is \textbf{not} the case. The best results are obtained when there is a max-pooling layer after each convolutional layer. Another important and related finding of our work is that the effect of max-pooling is reversed depending on how the dataset is split into training and testing sets. When the dataset is split into train and test sets in such a way that a human subject can appear only in one of them (this is called \textbf{subject-independent} split in the literature), max-pooling has a positive effect on performance. However, when the dataset is split purely randomly; that is, images from the same subject may appear in both train and test sets,  max-pooling has a negative effect. Although each image in the dataset is numerically different, \textbf{random} split transform the FER problem to a memorization problem. We validate this proposition with our empirical analysis (Table~\ref{tab:poolingResults}). Our findings raise three important questions:

\noindent \textbf{(1) Is information loss essential for generalization?} We show that, considering it as a tool for information loss where numerically dominant values suppress the others, max-pooling improves the classification performance. However, when the problem is transformed into a memorization challenge, having no max-pooling layers yields the best results.

\noindent \textbf{(2) Is a smaller model more open to teacher's supervision?} We show that, compared to training from scratch, KD becomes more effective as the network gets smaller. 
To the best of our knowledge, this is a novel finding
Whether this effect is specific to the FER problem is yet to be seen.

\noindent \textbf{(3) Why does the classification accuracy fluctuate as the \textit{temperature} hyperparameter is changed?} We show that, regardless of how the dataset is split (random or subject-independent), the accuracy fluctuates between low and high values as the temperature is swept across a wide range.


\section{Related work}
\label{relatedworks}

\subsection{Facial expression recognition (FER)}
We categorize the previous work as image (or frame) based and sequence-based. While image-based methods analyze individual images independently, sequence-based methods exploit the spatio-temporal information between frames. 

\paragraph{Image based.} There are three groups of work. Models that use 1) hand-crafted features (HCFs), 2) deep representations, and 3) both. Our work falls into the second group.

We do not focus on HCF models \cite{bartlett2005recognizing,feng2007facial,sikka2012exploring,zhong2012learning} here because they are obsolete (with the emergence of deep models) and in general, they do not achieve competitive results. 

Deep representations learned from face images are the main ingredients of ~\cite{liu2014facial,mollahosseini2016going,zhao2016peak,ding2017facenet2expnet,kim2017deep}. Liu et al. ~\cite{liu2014facial} proposed a loopy boosted deep belief network framework for feature learning, then used them in an AdaBoost classifier. Mollahosseini et al. \cite{mollahosseini2016going} introduced an inception network for FER. Their model is much larger compared to ours considering the two large fully connected layers at the end of their network. Zhao et al. \cite{zhao2016peak} proposed a peak-piloted GoogLeNet \cite{szegedy2015going} model which uses both peak and non-peak expression images during training. Training peak and non-peak images in pairs naturally requires their proposed back-propagation algorithm which adds complexity to implementation compared to our work. FN2EN  \cite{ding2017facenet2expnet} employs a multi-staged model production for FER. First, they train convolutional layers by mimicking \cite{ba2014deep} a pre-trained FaceNet \cite{parkhi2015deep}. Then, they append a $fc$ layer to the model for retraining. Recently, Kim et al. \cite{kim2017deep} introduced a deep generative contrastive model for FER. They combined encoder-decoder networks and CNNs into a unified network that simultaneously learns to generate, compare, and classify samples on a dataset. 

Finally, \cite{levi2015emotion} form a hybrid approach. They train CNNs with both the original input images and 3D mappings of local binary patterns \cite{ojala1996comparative}, then finalize via fine-tuning.

\paragraph{Sequence-based.} We can categorize sequence-based facial expression classifiers in the same three groups as in the case of image-based classifiers.  We do not focus on HCF based sequence models~\cite{guo2012dynamic,ghimire2013geometric,sanin2013spatio,sikka2016lomo} for the same reasons with the image-based case.

Liu et al.  \cite{liu2014learning} proposed manifold modeling of videos based on representations gathered via learned spatio-temporal filters. Kahou et al. \cite{ebrahimi2015recurrent} fused CNNs with recurrent neural networks (RNNs). CNN is used on static images to gather high-level representations which are then used by the RNN training.    Jung et al. \cite{jung2015joint} proposed a hybrid approach via two deep models. First, a 3D-CNN to extract the temporal appearance features from image sequences. Second, a fully connected model which captures geometrical information about the motion of the facial landmark points.

\subsection{Model size reduction}
The knowledge distillation (KD) method, which is the first and the most popular teacher-student style compression method,  \cite{hinton2015distilling} is described in  Section \ref{methodology}. Ba and Caruana \cite{ba2014deep} proposed using the $L_2$ loss on the logits to mimic the teacher (without any temperature hyperparameter). Both ideas are combined in FitNets \cite{romero2014fitnets}: they use the $L_2$ loss on logits in the pre-training phase for better initialization, then, they train the whole student network using the KD method. In contrast to FitNets, we choose a model that is much shallower than the teacher and avoid any pre-training of the student to prevent increasing the complexity of the overall training procedure. Iandola et al. \cite{iandola2016squeezenet} (SqueezeNet) proposed a CNN with no fully connected layers to reduce the model size, and preserved the classification performance via their fire modules.

\section{Methodology}
\label{methodology}
\subsection{Knowledge distillation}
\label{kd}
Formally, let $p_{t}$ be the softened output of the teacher's softmax, $z_{i}$ be the logits of the teacher, $p_{s}$ be the hard and $p'_{s}$ be the soft output of the student's softmax, $v_{i}$ be the logits of the student, $\lambda$ be the weight of distillation, $y$ be the ground truth labels, $N$ be the batch size, $T$ temperature and function $\mathcal{H}$ refers to the cross-entropy. Then:
\begin{equation}
\label{softenedSoftmax}
    p_{t} = \frac{e^{z_i / T}}{\sum _j e^{z_j / T}}, \quad p'_{s} = \frac{e^{v_i / T}}{\sum _j e^{v_j / T}}, \quad p_{s} = \frac{e^{v_i}}{\sum _j e^{v_j}}
\end{equation}

\noindent and the loss becomes
\begin{equation}
\label{lossFunc}
    \mathcal{L} = \lambda (\frac{1}{N} \sum^N_{n=1} \mathcal{H}(p_t, p'_s)) + (1 - \lambda) (\frac{1}{N} \sum^N_{n=1} \mathcal{H}(y, p_s)).
\end{equation}

\subsection{Network architectures}
\label{arch}
\paragraph{Teacher network} 
We use the Inception\_v3~\cite{szegedy2016rethinking} network as the teacher for its proven record of success on classification tasks~\cite{russakovsky2015imagenet}.

\paragraph{Student network} Our student network has a very simple architecture: two convolutional layers (conv1, conv2) and two fully-connected layers (fc1, fc2). We used rectified linear units (ReLU)~\cite{nair2010rectified} as activation functions. There are max-pooling layers after each conv layer (analysis on pooling vs. no-pooling is given in Section \ref{poolingAnalysis}). We created four different versions of the student network (namely M, S, XS, XXS) by varying the number of neurons at fc1. Table \ref{modelDetails} presents the sizes of these networks. We trained them using the KD \cite{hinton2015distilling} method. We compare their classification performances in sections \ref{ck} and  \ref{OuluCASIA}, speeds \& memory requirements in Section \ref{sizeAndSpeedAnalysis}.

\begin{table}
\caption{Sizes of the student networks.  All student networks have 4 layers: 2 convolutional layers (\textit{conv1} and \textit{conv2}) followed by 2 fully connected layers (\textit{fc1} and \textit{fc2}).  \textit{conv1} layer has 16 8x8 filters with stride 4,  \textit{conv2} layer has 32 4x4 filters with stride 2. \textit{fc2} layer has $8$ neurons. Different student networks differ only in the number of neurons in the \textit{fc1} layer.}
\label{modelDetails}
\centering
\begin{tabular}{ccc}
\toprule
Model &
\begin{tabular}{c}
     \# of  neurons \\ 
     in \textit{fc1} \\
\end{tabular} & 
\begin{tabular}{c}
     total \# of  \\ 
     parameters \\
\end{tabular} \\
\midrule
M 	&	256	&	900920	\\
S 	&	64	&	232184	\\
XS 	&	32	&	120728	\\
XXS	&	16	&	65000	\\
\bottomrule
\end{tabular}
\end{table}

\subsection{Implementation}
\label{implementation}
\paragraph{CK+ \& Oulu-CASIA} For each image in CK+, we apply the Viola Jones~\cite{viola2004robust} face detector, and for each image in Oulu-CASIA we use the already cropped versions. All images are converted to grayscale. Then, in order to augment the data, we extract $8$ crops ($4$ from each corner and $4$ from each side) from an image with dimensions of $84$x$84$ for students and $256$x$256$ for the teacher. There is no difference on hyperparameter selections for the trainings on  CK+ and Oulu-CASIA. As done in previous work, we report the average 10-fold cross-validation (CV) performance. For both the teacher and students, trainings are finalized after $3000$ epochs.

\paragraph{Teacher Network} We employ a Inception\_v3 network trained on ImageNet \cite{russakovsky2015imagenet}, and fine-tune it on FER datasets. The base learning rate is set as $10^{-4}$ and remained constant through iterations, mini-batch size is $64$, and the learning method is Adam~\cite{kingma2014adam}.

\paragraph{Vanilla \& Student Networks} We have the same hyperparameters across all of the different model sizes for both vanilla and student trainings. ``Vanilla'' training means that the network is trained from scratch without any teacher guidance.  Weights and biases are initialized using Xavier initialization \cite{glorot2010understanding}. Network architectures are implemented using Tensorflow \cite{abadi2016tensorflow}. We used Adam~\cite{kingma2014adam} with a learning rate of $10^{-4}$. The dropout~\cite{srivastava2014dropout} rate is $0.5$, mini-batch is $64$ and the weight of the distillation $\lambda$ is $0.5$ (see Section \ref{kd}) for all student models. Selected model sizes are $900$K, $232$K, $121$K and $65$K parameters respectively, which are produced by decreasing the size of the $fc1$ layer (see Table \ref{modelDetails}). Training operations are finalized after $3000$ epochs for all models and the XXS student model is denoted as \textbf{MicroExpNet}. Empirical results are given in Table \ref{tab:classificationCK_Oulu}; note that for student networks we only list the best performers across different temperatures (selected using cross-validation). Furthermore, student models are used in temperature selection tests (for detailed explanation see Section \ref{temperatureAnalysis}). The results we report for these models are obtained by averaging the 10-fold cross-validation performances.


\begin{table}
	\caption{The effect of max-pooling. Classification performances of the candidate student models for 1000 epochs of training. $p1$ indicates that there is only one max pooling layer after conv1, $p2$ indicates that there is only one max pooling layer after conv2, $p12$ indicates that each convolution layer is followed by a max-pooling layer, and $v$ indicates that there is no pooling layer at all. The smaller the network, the more max-pooling degrades the performance for \textbf{random split} whereas the opposite holds for \textbf{subject-independent split}.}
	\label{tab:poolingResults}
	\centering
	\resizebox{.5\textwidth}{!}{\begin{tabular}{c|lcc|lcc}
	 
		\toprule
		& Model & CK+ & Oulu-CASIA & Model & CK+ & Oulu-CASIA \\
		\midrule
		\parbox[t]{2mm}{\multirow{8.5}{*}{\rotatebox[origin=c]{90}{Random}}}  
		& $v_{M}$     	  		  & $97.93$\%   		& $97.68$\%  & 			\textbf{$v_{XS}$}  & $\textbf{93.41}$\% & $\textbf{88.73}$\%
		\\

		& \textbf{$p1_{M}$}      & $\textbf{97.99}$\%  & $\textbf{97.79}$\% & 	$p1_{XS}$          & $91.85$\%          & $80.16$\% \\
		& $p2_{M}$               & $97.41$\%           & $96.64$\%  & 			$p2_{XS}$          & $86.84$\%          & $77.88$\% \\
		& $p12_{M}$              & $97.39$\%           & $97.47$\%  &			$p12_{XS}$         & $88.07$\%          & $77.04$\% \\

		\cmidrule{2-7}
		& $v_{S}$     	  		  & $96.65$\%   		& $92.95$\%  &			\textbf{$v_{XXS}$} & $\textbf{81.91}$\% & $\textbf{73.64}$\% \\
		& \textbf{$p1_{S}$}      & $\textbf{96.73}$\%  & $\textbf{93.22}$\% & 	$p1_{XXS}$         & $69.05$\%          & $52.99$\% \\
		& $p2_{S}$               & $94.09$\%           & $88.61$\%  & 			$p2_{XXS}$         & $77.74$\%          & $66.84$\% \\
		& $p12_{S}$              & $94.39$\%           & $88.72$\%  &			$p12_{XXS}$        & $78.52$\%          & $61.71$\% \\
		\bottomrule
		\parbox[t]{2mm}{\multirow{8.5}{*}{\rotatebox[origin=c]{90}{Subject-independent}}} &
		$v_{M}$     	  		  & $81.23$\%   		& $60.87$\%  & 			$v_{XS}$  & $77.14$\% & $53.73$\%  
		\\

		& \textbf{$p1_{M}$}      & $\textbf{81.57}$\%  & $\textbf{62.46}$\% & 	$p1_{XS}$          & $77.14$\%          & $53.41$\% \\
		& $p2_{M}$               & $78.77$\%           & $60.21$\%  & 			$p2_{XS}$          & $78.42$\%          & $57.51$\% \\
		& $p12_{M}$              & $79.95$\%           & $60.53$\%  &			\textbf{$p12_{XS}$}         & $\textbf{79.78}$\%          & $\textbf{57.54}$\% \\
		
		\cmidrule{2-7}
		& $v_{S}$     	  		  & $79.73$\%   		& $58.18$\%  &			$v_{XXS}$ & $71.36$\% & $44.33$\% \\
		& \textbf{$p1_{S}$}      & $\textbf{81.25}$\%  & $\textbf{59.49}$\% & 	$p1_{XXS}$         & $67.04$\%          & $34.04$\% \\
		& $p2_{S}$               & $78.75$\%           & $57.37$\%  & 			$p2_{XXS}$         & $76.91$\%          & $54.62$\% \\
		& $p12_{S}$              & $79.71$\%           & $57.25$\%  &			\textbf{$p12_{XXS}$}        & $\textbf{78.44}$\%          & $\textbf{55.03}$\% \\
		\bottomrule
		\end{tabular}}
	\vspace{-3pt}
\end{table}

\section{Experiments}
\label{experiments}

\subsection{Max. Pooling vs. No Pooling Analysis}
\label{poolingAnalysis}

Facial expressions are located mostly on eyes and mouth ~\cite{ekman1997face}, and they form only a small fraction of a frontal face image. The idea is to capture these subtle indicators of emotion by preserving the pixel information across layers. Therefore, our starting point was a CNN with no pooling layers. However, in order to validate our intuition, we build three variations containing max-pooling layers for each student. All pooling layers have $2$x$2$ filters with stride $2$. All hyperparameters mentioned at Section \ref{implementation} apply to these variations as well. We call them candidate expression networks. These candidates are explained in Table \ref{tab:poolingResults}.  

From the results in Table \ref{tab:poolingResults}, we draw the following conclusions. When models are large enough, the network capacity for learning dominates pooling effects. For instance, for the size M, classification performances of candidates are very close to each other. For size S, poolings in later layers drop the performance but early pooling is still the most profitable. After this point (XS and XXS), we begin to see an interesting difference between the results of random and subject-independent split experiments. \textbf{For random split, we see the advantage of not having any pooling layers with significant gains in performance.} Since the trained candidates see the same subjects in both training and test (for $\approx 80\%$ of the subjects), although the images are numerically different, we think that the resemblance transforms the FER problem to a memorization challenge. Hence, the information loss caused by the pooling layers drops the performance. \textbf{On the contrary, for subject-independent split where test subjects are not seen during the training, we observe the advantage of having pooling layers.} Thus, the intuition mentioned above does not seem to hold. It is also interesting to observe that the second pooling layer seems to be a much critical point of improvement than the first pooling layer. Nevertheless, combining these observations with our intention to reduce the model size, we decided to employ the architecture with two pooling layers as the foundation of our student networks.

Note that adding a pooling layer drops the number of parameters; thus, prevents a proper performance comparison. Therefore, we did two modifications to increase the model size, in order to make it a fair comparison. First, when we add a pooling layer after the first convolutional layer, we decrease the stride of the first conv layer from $4$ to $2$. This directly recovers all parameters that has been lost. Second, when we add a pooling layers after the second convolutional layer, we increase the number of outputs of the first fully connected layer by $3$-fold. This results in having slightly less parameters than the original one (${v}$). 

\begin{table*} 
\caption{The number of images per  expression class in Ck+ and Oulu-CASIA.}
\label{tab:datasets}
\centering
 \begin{tabular}{lccccccccc} 
 & Anger & Contempt & Disgust & Fear & Happy & Sad & Surprise & Neutral & All \\ 
 \hline
 CK+ & 135 & 54 & 177 & 75 & 207 & 84 & 249 & 593 & 1574   \\ 
 Oulu-CASIA & 240 & - & 240 & 240 & 240 & 240 & 240 & - & 1440  \\  
 \hline
 \end{tabular}
\end{table*}

\subsection{The CK+ dataset}
\label{ck}
CK+ is a widely used benchmark database for facial expression recognition. This database is composed of $327$ image sequences with eight emotion labels: anger, contempt, disgust, fear, happiness, sadness, surprise and neutral. There are $123$ subjects. As done in previous work, we extract the last three and the first frames of each expression sequence when images are labeled. When unlabeled, we only extracted first frames as neutral. The total number of images is $1574$ (see at Table \ref{tab:datasets}), which is split into $10$ folds. 

\paragraph{Training in Isolation} We evaluate the pre-trained Inception\_v3 via fine-tuning on CK+. Then, we train four models, namely VanillaExpNet$_{M}$, VanillaExpNet$_{S}$, VanillaExpNet$_{XS}$, and VanillaExpNet$_{XXS}$, from scratch. At this stage, we did not employ knowledge distillation. For all models, we used $3000$ epochs for training, and the classification performances are shown in Table \ref{tab:classificationCK_Oulu}. Although Zhao et al. ~\cite{zhao2016peak} seem to achieve better performance than Inception\_v3 (in Table \ref{tab:classificationCK_Oulu}), they use only $6$ emotion categories, whereas we use all of the $8$ emotion categories. In the light of these results, we chose Inception\_v3 as the teacher for the knowledge distillation stage.

\begin{figure}
\centering
\includegraphics[scale=0.45]{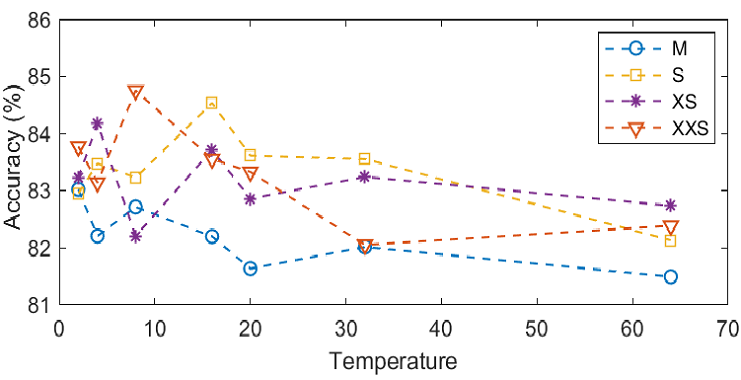}
\caption{Classification performances of the student networks across different temperatures on the CK+ dataset using \textbf{subject-independent splits}.}
\label{fig:ck_temperature}
\end{figure}

\begin{figure}
\centering
\includegraphics[scale=0.45]{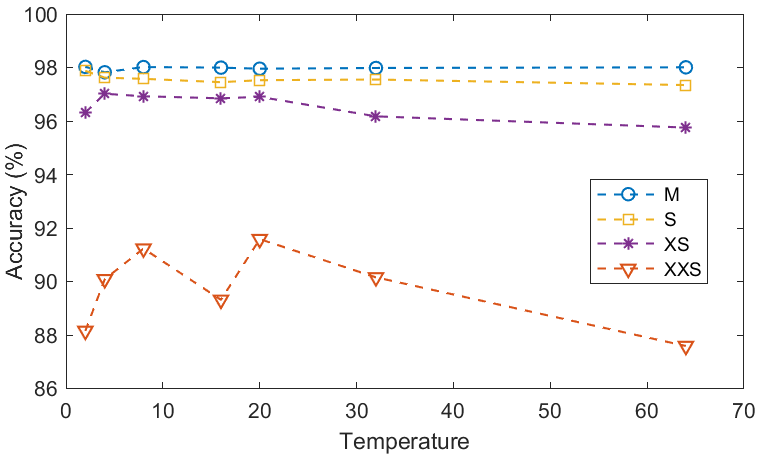}
\caption{Classification performances of the student networks across different temperatures on the CK+ dataset using \textbf{random splits}.}
\label{fig:ck_temperature_random}
\end{figure}

\paragraph{Training with Supervision} We evaluate four students, namely StudentExpNet$_{M}$, StudentExpNet$_{S}$, StudentExpNet$_{XS}$, and StudentExpNet$_{XXS}$, via knowledge distillation on CK+. At this stage, we use the teacher's supervision to improve the learning. As explained in Section \ref{implementation}, we need to tune the \textit{temperature} for each student since it is considered to be  correlated with model size. Therefore, we conducted an extensive experiment on classification performances for a wide range of $temperatures$. The results are reported in Figure \ref{fig:ck_temperature}. According to these results, fluctuations between performances are increased while models are getting smaller. Consequently, it suggests that large networks are more tolerant to the changes in the temperature. This observation also holds for the random split case as shown in Figure \ref{fig:ck_temperature_random}. 

Best performers, based on their average classification performances for $10$-fold cross-validation, across different temperatures are then used for performance comparison in Table \ref{tab:classificationCK_Oulu}. Our findings show that KD can be used to gain back some of the performance lost by decreasing the model size. 

\begin{table}
\caption{Average classification accuricies (over 10-folds) of different methods on CK+ and Oulu-CASIA datasets.}
\label{tab:classificationCK_Oulu}
\centering
\begin{tabular}{l|cc|c}
\toprule
\multirow{2}{*}{Method} 	& \multicolumn{2}{c|}{CK+}    & 
\multirow{2}{*}{
\begin{tabular}{c}Oulu-\\CASIA\end{tabular}
} \\\cmidrule{2-3}
		& 6 cls & 8 cls & \\
\midrule
CSPL~\cite{zhong2012learning}                		& 89.9\%  & -		& - \\
3DCNN-DAP~\cite{liu2014deeply}                		& 92.4\%  & -		& - \\
Inception~\cite{mollahosseini2016going}				& 93.2\%  & -		& - \\
AdaGabor~\cite{bartlett2005recognizing}				& 93.3\%  & -		& - \\
AdaLBP~\cite{zhao2011facial}				        & - 	  & - 		& 73.54\% \\
STM-ExpLet~\cite{liu2014learning}					& 94.2\%  & -		& 74.59\% \\
Atlases~\cite{guo2012dynamic}                       & - 	  & -	 	& 75.52\% \\
LOMo~\cite{sikka2016lomo}							& 95.1\%  & -		& 82.10\% \\
LBPSVM~\cite{feng2007facial}						& 95.1\%  & -		& - \\
BDBN~\cite{liu2014facial}							& 96.7\%  & -		& - \\
DTAGN~\cite{jung2015joint}							& 97.3\%  & -		& 81.46\% \\
FN2EN~\cite{ding2017facenet2expnet}					& 98.6\%  & 96.8\% 	& 87.71\% \\
DCN~\cite{zhang2016facial}                          & 98.9\%  & -		& \\                      
PPDN~\cite{zhao2016peak}							& 99.3\%  & -		& 84.59\% \\
AU-Aware~\cite{liu2013aware}						& - 	  & 92.1\%  & \\
GCNet~\cite{kim2017deep}                            & - 	  & 97.3\%  & 86.39\% \\
\textbf{TeacherExpNet} 								& - & \textbf{97.6\%} & \textbf{85.83\%} \\
VanillaExpNet$_{M}$                                 & - & 78.8\% & 56.81\% \\
VanillaExpNet$_{S}$                                 & - & 78.6\% & 55.53\% \\
VanillaExpNet$_{XS}$                                & - & 77.2\% & 54.67\% \\
VanillaExpNet$_{XXS}$                               & - & 75.3\% & 56.71\% \\
StudentExpNet$_{M}$                                 & - & 83.1\% & 63.81\% \\
StudentExpNet$_{S}$                                 & - & 83.6\% & 62.01\% \\
StudentExpNet$_{XS}$                                & - & 83.7\% & 61.76\% \\
\textbf{MicroExpNet}       							& - & \textbf{84.8\%} & \textbf{62.69\%} \\
\bottomrule
\end{tabular}
\end{table}

\subsection{The Oulu-CASIA dataset}
\label{OuluCASIA}
Oulu-CASIA has 480 image sequences taken under \textit{dark, strong, weak} illumination conditions. In this experiment, as also done in previous work, we used only videos with \textit{strong} condition captured by a VIS camera. In total, there are 80 subjects and six expressions: anger, disgust, fear, happiness, sadness, and surprise. Similar to CK+, the first frame is always neutral while the last frame has the peak expression. All studies we have encountered on Oulu-CASIA database use only the last three frames of the sequences, so we also use the same frames. Therefore, the total number of images is 1440. As in the earlier studies, a 10 fold CV is performed, and the split is subject independent. 

\paragraph{Training in isolation} The same approach taken for CK+ is employed for Oulu-CASIA. The classification performances are shown in Table \ref{tab:classificationCK_Oulu}. According to the table, Inception\_v3 performs on par with the state-of-the-art solutions whereas our vanilla models failed to achieve competitive results.

\begin{figure}
\centering
\includegraphics[scale=0.45]{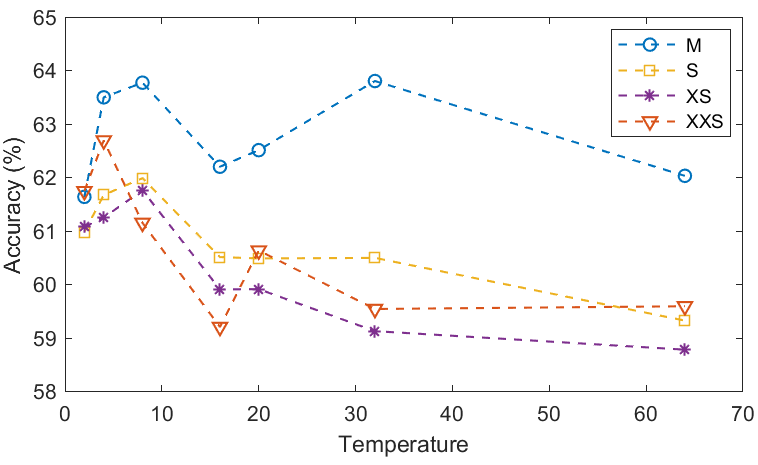}
\caption{Classification performances of the student networks across different temperatures on the Oulu-CASIA dataset using \textbf{subject-independent splits}.}
\label{fig:oulu_temperature}
\end{figure}

\begin{figure}
\centering
\includegraphics[scale=0.45]{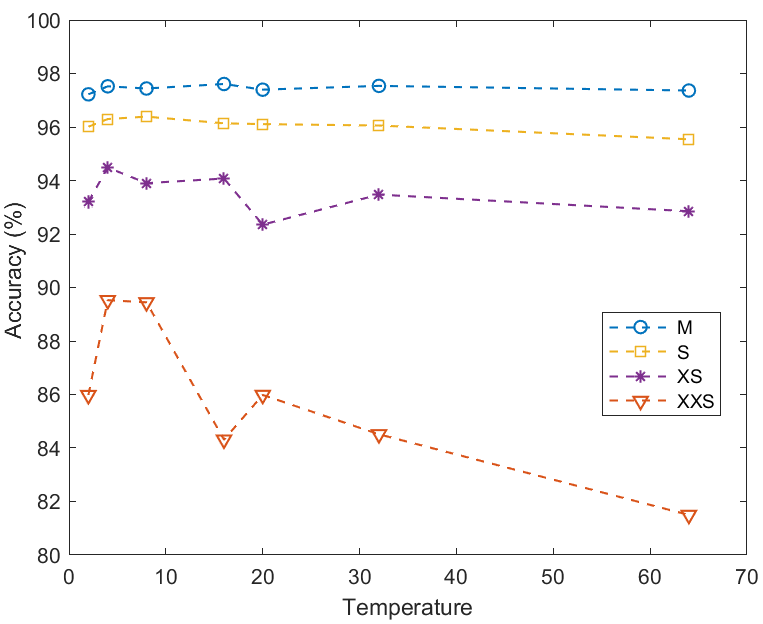}
\caption{Classification performances of the student networks across different temperatures on the Oulu-CASIA dataset using \textbf{random splits}.}
\label{fig:oulu_temperature_random}
\end{figure}

\paragraph{Training with supervision} The same explanations on students for CK+ also apply to Oulu-CASIA experiments. The results are reported in Figure \ref{fig:oulu_temperature} from which, we can observe a similar fluctuating behavior as seen in the CK+ experiments. Once again, we can see that large networks are more tolerant to the changes in the temperature than the smaller ones. In addition, as in CK+ experiments, this observation also holds for the random split case as shown in Figure \ref{fig:oulu_temperature_random}. 

Best performers across different temperatures are then used for performance comparison in Table \ref{tab:classificationCK_Oulu}. We can still observe that the student models perform better than vanilla models (which are trained from scratch without any teacher supervision) for facial expression recognition. 

\subsection{Temperature analysis}
\label{temperatureAnalysis}
Temperature is a hyperparameter to control the uncertainty in teacher's output. This uncertainty may be used as similarity information between different classes to enhance the training. However, there is no formulation for selecting the most effective temperature; it is set empirically. Hence, we did a grid search for temperatures of [2, 4, 8, 16, 20, 32, 64] with 10-fold cross-validation across all of our student networks using both CK+ (see Figure \ref{fig:ck_temperature}) and Oulu-CASIA (see Figure \ref{fig:oulu_temperature}) datasets using a subject-independent train \& validation split. Moreover, we did a grid search for a random train \& validation split as well (see Figures \ref{fig:ck_temperature_random} and \ref{fig:oulu_temperature_random}).

According to the results, smaller models are more prone to temperature changes in general, and performances for a given temperature seem rather stochastic. However, large models show different characteristics for the random split case and subject-independent split case. When the subject-independent split is used, we observe fluctuations in performance for all models regardless of their size. Whereas for the random split case, large models have relatively stable performances. Nevertheless, when calibrated adequately, KD improves the overall FER performance for the subject-independent split case.

\subsection{Model size and speed analysis}
\label{sizeAndSpeedAnalysis}
We show the comparison of the model sizes in megabytes in Table \ref{tab:modelSizesAndSpeeds}. Our smallest FER model MicroExpNet takes less than $1$ MB to store which is $100$x smaller than our teacher network (Inception\_v3), and, it has $335$x fewer parameters than the teacher. We also show the running times to process one image in milliseconds (average of 1000 runs) in Table \ref{tab:modelSizesAndSpeeds}. According to the table, MicroExpNet achieves the best performance by classifying the facial expression in an image in less than $1$ ms on an Intel i7 CPU. Ultimately, when compared to its teacher, MicroExpNet is $234$x faster on Intel i7, and $85$x faster on GTX1050.

\setlength{\tabcolsep}{3pt}
\begin{table}
\begin{center}
\caption{Memory requirements and average per-image running times of different FER models.}
\label{tab:modelSizesAndSpeeds}
\vspace{-13pt}
\resizebox{.5\textwidth}{!}{
\begin{tabular}{lrrrrr}
\\
\toprule
Model & \begin{tabular}{c}\# of \\ params \end{tabular} & \begin{tabular}{c}Size \\(MB) \end{tabular} & i7-7700HQ & GTX1050 & Tesla K40\\
\midrule
TeacherExpNet						&$21.8$M					&  $88.13$  & $124.22$ ms  &  $83.25$ ms &           -  \\
FN2EN~\cite{ding2017facenet2expnet} &$11$M						&  $42.42$  &  $96.08$ ms  &  $23.81$ ms &  $13.09$ ms  \\
PPDN~\cite{zhao2016peak}			&$6$M						&  $23.93$  &  $57.18$ ms  &   $9.12$ ms &  $13.11$ ms  \\
StudentExpNet$_{M}$                 &$900$K             		&  $10.88$  &   $0.89$ ms  &   $1.13$ ms &   $1.74$ ms  \\
StudentExpNet$_{S}$                 &$232$K           			&   $2.91$  &   $0.78$ ms  &   $1.08$ ms &   $1.69$ ms  \\
StudentExpNet$_{XS}$                &$121$K           			&   $1.52$  &   $0.63$ ms  &   $0.97$ ms &   $1.63$ ms  \\
\textbf{MicroExpNet}     			&$\textbf{65}$\textbf{K}   	& $\textbf{0.88}$ & $\textbf{0.53}$ \textbf{ms} & $\textbf{0.97}$ \textbf{ms} & $\textbf{1.52}$ \textbf{ms}\\
\bottomrule
\end{tabular}
}
\end{center}
\end{table}

\section{Conclusion}
\label{conclusion}
We presented an extensive analysis of the creation of a microarchitecture, called the MicroExpNet, for facial expression recognition (FER) from frontal face images.

From our experimental work, we have drawn the following conclusions. (1) Information loss achieved via max-pooling and the KD method both improve the performance especially when the network is small, (2) we showed that a simple change in the approach taken for the separation of train \& validation sets results in drastic changes in the problem definition, and thus in the performance observations, (3) KD's effect gets more prominent as the network size decreases. Whether this effect is generalizable to other problems/datasets is yet to be seen in future work. (4) The temperature hyperparameter (in KD) should be tuned carefully for optimal performance. Especially when the network is small, the final performance fluctuates with temperature. 


\bibliographystyle{IEEEtran}  
\bibliography{IEEEfull,mybibfile}
\vspace{12pt}

\end{document}